\documentclass[a4paper, 10pt, conference]{ieeeconf}      
\usepackage{FG2025}

\FGfinalcopy 

\usepackage{graphicx}
\usepackage{amsmath}
\usepackage{amssymb}
\usepackage{booktabs}       
\usepackage{amsfonts}       
\usepackage{nicefrac}       
\usepackage{microtype}      
\usepackage{xcolor}         
\usepackage{epsfig}
\usepackage{caption}
\usepackage{dsfont}
\usepackage{subcaption}
\usepackage{multirow}
\usepackage{bm}
\usepackage{balance}
\usepackage{soul,color}
\usepackage[ruled]{algorithm2e}
\usepackage[symbol]{footmisc}
\newcommand\xleftrightarrow[1]{%
  \mathbin{\ooalign{$\,\xrightarrow{#1}$\cr$\xleftarrow{\hphantom{#1}}\,$}}}
\usepackage{xcolor}

\def\FGPaperID{235} 

\title{\LARGE \bf
FaceCloak: Learning to Protect Face Templates
}

\author{\parbox{16cm}{\centering
   {\large Sudipta Banerjee, Anubhav Jain, Chinmay Hegde and Nasir Memon}\\
   {\normalsize
   New York University\\}}
}

\usepackage{fancyhdr}

\usepackage[pagebackref=true,breaklinks=true,letterpaper=true,colorlinks,bookmarks=false]{hyperref}

\begin{document}

\maketitle

\thispagestyle{fancy}

\begin{abstract}

Generative models can reconstruct face images from encoded representations (templates) bearing remarkable likeness to the original face, raising security and privacy concerns. We present \textsc{FaceCloak}, a neural network framework that protects face templates by generating smart, renewable binary cloaks. Our method proactively thwarts inversion attacks by cloaking face templates with unique disruptors synthesized from a single face template on the fly while provably retaining biometric utility and unlinkability. Our cloaked templates can suppress sensitive attributes while generalizing to novel feature extraction schemes and outperform leading baselines in terms of biometric matching and resiliency to reconstruction attacks. \textsc{FaceCloak}-based matching is extremely fast (inference time =0.28 ms) and light (0.57 MB). We have released our \href{https://github.com/sudban3089/FaceCloak.git}{code} for reproducible research.

\end{abstract}


\section{INTRODUCTION}\label{sec:Intro}

\textbf{Motivation.} Face recognition systems (FRS) are susceptible to various forms of attacks by motivated adversaries~\cite{BTS_1, BTS_2, handbook}. Template inversion attacks attempt to invert a given face template, typically produced using a deep neural feature extraction model, to recover its original face image~\cite{masq, nbnet, Ahmad2022, dong2023}. In order to carry out such an attack, the adversary requires access to the target template and either white-box or black-box access to the feature extractor model. An inverted template can pose multiple concerns: the adversary can then use the original biometric sample to generate system-specific templates for unauthorized access to multiple services (banking, phone, medical records, etc.). In addition, the adversary can deduce protected attributes such as the gender, age, and ethnicity of the target, raising privacy concerns. Some well-known face template inversion methods that use generative models are NbNet, StyleGAN-inversion and Arc2Face~\cite{nbnet, GANInv, Arc2face}. Template-level attacks are not new, but lately, novel attacks are emerging with higher diversity and increased attack success due to three factors: 
{(i) rapid surge of IoT devices connecting users to \textit{remote applications} (mobile banking, online interview)};
{(ii) widespread deployment of \textit{biometric authentication} for easy access (Apple FaceID, Google Pay using Face Unlock)};
{(iii) a gamut of \textit{open-source generative tools} that can instantly simulate realistic looking synthetic media, thus attracting the attention of malicious agents (diffusion models, LLMs).}

As a result, biometric template protection schemes are recommended security measures in real-world FRS deployments~\cite{BTPSurvey}. Most template protection schemes are based on semantic feature transformations of the given template~\cite{Trans_1, Trans_2, polyprotect}, or cryptographic encryption schemes~\cite{HE_1, HE_2, hers, BTASHE}, or hybrid approaches. However, all such schemes suffer from performance-security trade-offs or can leak sensitive attributes~\cite{Poly_FE}.

\begin{figure*}[t]
    \begin{subfigure}[b]{0.625\textwidth}
         \includegraphics[width=\textwidth]{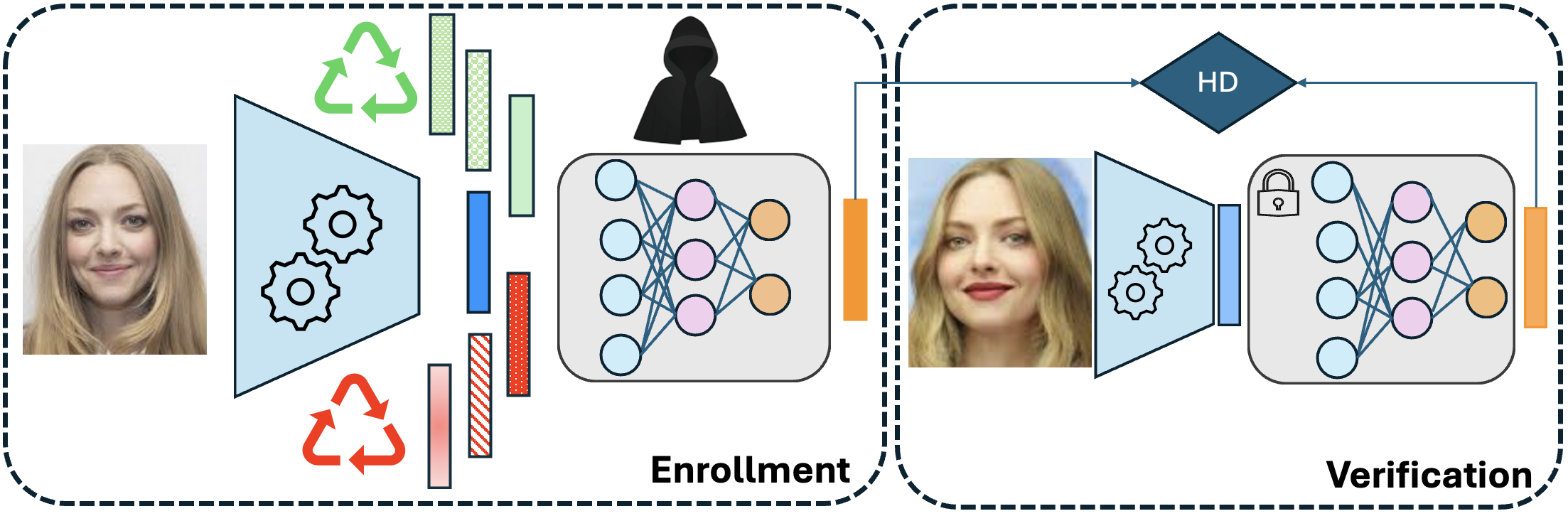}
         \caption{}
     \end{subfigure} \hfill
    \begin{subfigure}[b]{0.35\textwidth}
         \includegraphics[width=\textwidth]{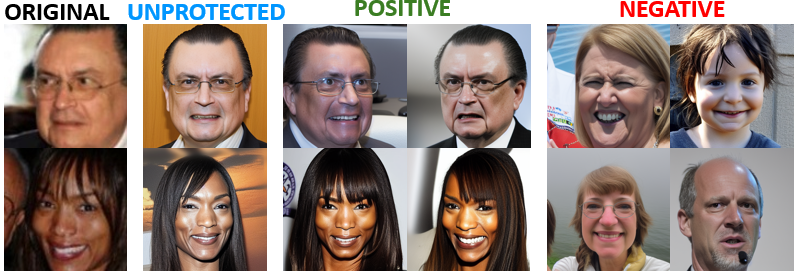}
         \caption{}
     \end{subfigure}
    \caption{ \textbf{(a)} \textsc{FaceCloak} framework. For an input image, a pre-trained face template extractor produces the \textcolor{blue}{unprotected} face template. During enrollment, we first produce $k$ \textcolor{green}{positive} and $k$ \textcolor{red}{negative} disruptors from the \textit{single} face template and then train our lightweight network with the $2k+1$ (including the original template) inputs supervised using biometric identity, binarization and diversity loses resulting in \textcolor{orange}{protected} enrolled template. During verification, we pass the query template through the trained \textsc{FaceCloak} and compare with the enrolled template using Hamming distance (HD). Our method does not require user-specific seed or keys and can produce renewable disruptors, resulting in an automatically secure randomized network with biometric retention. \textbf{(b)} Examples of original images from the LFW dataset, the inverted outputs of their respective unprotected templates and the corresponding inverted outputs of positive (noise and mask), and negative (orthogonalization and synthetic) disruptors used in \textsc{FaceCloak}. We used Arc2Face~\cite{Arc2face} to perform the inversion.}
    \label{fig:outline}
\end{figure*}

\textbf{Our approach.} In this work, we propose \textsc{FaceCloak}, a novel framework that protects a given face template in a manner that is resistant to face reconstruction. \textsc{FaceCloak} strategically {learns} to combine a set of disruptor templates to shield the original underlying template while preserving biometric utility. Intuitively, we formulate the problem of generating the protected template as a binary vector for \textit{each} unprotected face template, similar to generating deep hash codes for image retrieval~\cite{DHN}. This requires learning a \textit{unique} hash (binary vector) which will be computationally difficult to invert (mitigate inversion) while maintaining intra-class variations (biometric retention). 

\textsc{FaceCloak} uses a neural network that takes a single \textcolor{black}{unprotected} face template as input, then derives a set of disruptors consisting of \textcolor{black}{positive} (noise and mask), and \textcolor{black}{negative} instances (synthetic templates and orthogonalization); finally combines the original template with the disruptors to construct the \textcolor{black}{protected} or cloaked template supervised by biometric identity, binarization and diversity losses. Our network with two fully connected layers, batch normalization and activations is non-trivial to invert while the loss functions retain matching performance; see Fig~\ref{fig:outline}. \\
\noindent \textbf{Use case:} During enrollment, \textsc{FaceCloak} accepts any face template as input, and generates disruptors to create a protected template. The gallery will save the trained network and its output as the \textit{enrolled} template, but discard the disruptors. During verification, the \textit{query} template goes through the trained \textsc{FaceCloak}. The output is then compared with the enrolled protected template from the gallery using Hamming distance for decision. \textit{Our method does not require user-specific key management, is efficient and highly effective in ensuring biometric matching, security and privacy.}

\vspace{0.5pt}
\textbf{Contributions.} Our main contributions are as follows.
\begin{enumerate}
\item We propose \textsc{FaceCloak}, a neural network-based face template protection scheme that uniquely maps a single face template to a cloaked binary vector by learning to optimally combine a set of disruptors with the original template using biometric identity, binarization and diversity loss functions. \textsc{FaceCloak} is a light-weight network with an average size of \texttt{0.57 MB} (a trained face template extractor such as ArcFace is $\sim$130MB) with average time costs: enrollment time\texttt{=0.41 ms}, inference time\texttt{=0.28 ms} and training time\texttt{=1.54 secs} on a RTX8000 GPU. Template size in memory: \texttt{5 KB} (FaceNet512), \texttt{2 KB} (ArcFace) and \texttt{1 KB} (Ours) using 32 bit float representation.

\item \textsc{FaceCloak} satisfies all three criteria for a biometric template protection scheme: provably retains \textit{biometric matching}, exhibits strong \textit{irreversibility}, and enforces \textit{unlinkability}. We demonstrate the properties by evaluating on two datasets (LFW and CFP), and two face template extraction methods (ArcFace and FaceNet512).

\item Our method is generalizable to novel feature extraction schemes. For validation, we propose ArcFace-OPL (ArcFace finetuned with Orthogonal Projection Loss) for improving disruptor creation and test \textsc{FaceCloak} for robustness. We further demonstrate that our template protection scheme can inherently suppress sensitive demographic attributes, such as, gender.
\end{enumerate}

\vspace{0.5pt}
\section{PRIOR WORK}

\textit{Face template protection} comes under the purview of biometric template protection (BTP) which focuses on protecting the biometric templates to safeguard the privacy of the owner and at the same time provide security to the authentication framework. They focus on ensuring irreversibility of the protected templates, \textit{i.e.}, the protected templates will be resilient against inversion attacks,  retention of biometric performance, \textit{i.e.,} minimal loss of biometric utility, and unlinkability, \textit{i.e.,} a new protected template can be regenerated which will be unlinked to the previously compromised protected template. Classical approaches apply \textit{feature transformation} to convert original face templates to protected templates with the help of user-specific transformation functions~\cite{Trans_1, Trans_2, polyprotect}. Homomorphic encryption (HE) is one of the most popular face BTP schemes as it allows comparison in the encrypted domain while ideally causing no biometric performance degradation. However, computational overhead and the need for maintaining the decryption algorithm as secret pose disadvantages for HE. Currently, there has been significant progress in speeding up HE operations through quantization and dimensionality reduction~\cite{HE_1, BTASHE, hers, HE_2}. Fuzzy Commitment scheme~\cite{FC1, FC2} and Fuzzy Vault scheme~\cite{FV} apply cryptographic hash functions to randomly generated codewords, which are related to the original face templates via a mathematical function. Secure sketch for biometric templates utilized auxiliary user information and provides a bound on the performance loss in terms of relative entropy~\cite{SF}. The mapping between the face template and the codeword can be established via the Locality Sensitive Hashing (LSH) algorithm~\cite{lsh}, while IronMask~\cite{IronMask} defined the mapping in terms of a linear transformation represented by an orthogonal matrix. Secure Face~\cite{SecFace} uses a randomized CNN and user-specific key for template protection. A deep rank hashing (DRH) network and a cancellable identification scheme was proposed in~\cite{DRHN}. BioHashing~\cite{Biohashing} performed template protection via discretization of original template using a seed value and random number, and a secret key sharing scheme. MLP-hash~\cite{MLPhash} used a user-specific randomly weighted MLP to create binary protected templates. Index-of-Maximum (IoM)~\cite{IOM} hashing used ranking of LSH for template protection. Refer to~\cite{BTPSurvey} for a comprehensive survey. Recently, SLERPFace~\cite{SLERP} used spherical and linear interpolation of templates to rotate on the hypersphere to become noise. Existing methods either rely on user-specific parameters/keys~\cite{polyprotect, BTASHE}, or can be computationally prohibitive~\cite{MLPhash, IronMask} or maybe restricted to a specific type of feature extractor, such as SLERPFace~\cite{SLERP}, which requires unit normalized features based on angular margin, and does not analyze generalizability across other features such as FaceNet~\cite{facenet} or MagFace~\cite{magface}.

\vspace{0.5pt}
\section{PROPOSED METHOD}


\textsc{FaceCloak} consists of two steps: \textit{Disruptor Creation} and \textit{Cloak Generation}.

\noindent \textbf{(A) Disruptor Creation:} The underlying target template is unprotected and susceptible to inversion attacks. An effective way of shielding it is to transform the template strategically. To achieve this, we utilize ``disruptors'' that can transform the protected template while maintaining biometric retention and mitigating inversion or reconstruction. We create positive disruptors that are similar or highly correlated to the original template, and negative disruptors that are dissimilar or weakly correlated to the original template. The purpose of the positive disruptors is to simulate intra-class variations that will help with biometric retention, which is the limitation of existing neural network-based biometric protection schemes~\cite{SecFace, MLPhash}. To synthesize positive disruptors, we randomly add Gaussian noise with low variance and/or mask some of the elements in the original template. On the other hand, negative disruptors aim to perturb the original template to shield against inversion attacks. Negative disruptors are synthesized using orthogonalization and/or templates from synthetic faces. \\

\noindent \textbf{(B) Cloak Generation:} After creating the set of positive and negative disruptors, the next step is how to best combine them with the unprotected template to ensure (i) minimal loss in biometric matching between unprotected template and cloaked (protected) template, and (ii) maximal uniqueness in the cloaked templates. To achieve this, we use a combination of biometric identity, binarization and diversity losses.\\

\noindent \textbf{Methodology.} We describe the terminology and notations used in designing \textsc{FaceCloak}. 
\textbf{Input:} Face template, $\textbf{t} \in \mathbb{R}^{1\times512}$. 
\textbf{Output:} Binary cloak, $\textbf{h} \in \{-1,+1\}^d$, where $d$ is the hash size. 
We use a shallow neural network $f$, defined as $f: \textbf{h} = f(\textbf{t})$.
\textbf{Architecture:} The neural network $f$ consists of two linear layers ($512 \rightarrow 256 \rightarrow d$) each followed by batch normalization and we use ReLU as the first activation function and Tanh as the second activation function to ensure the outputs are mapped in the range $[-1,+1]$. We tried $d={32, 64, 128, 256}$, and empirically observed $d=64$ was optimal in terms of performance and training cost. 

\textbf{Training:} We accept a face template $\textbf{t}$ as input. We then create $k$ \textit{positive} disruptors by applying input perturbations such as (i) adding Gaussian noise with \texttt{mean=0, std=0.2} and (ii) applying a mask (sampling from Bernoulli distribution with masking probability \texttt{p=0.2}) to assign some elements of template, $\textbf{t}$ to zero. We create $k$ negative disruptors via (i) Gram Schmidt orthogonalization, and (ii) synthetic face templates from a generative model (we use StyleGAN3~\cite{StyleGAN3} with truncation factor $\texttt{psi}=0.7$). Thus, we have a total of $2k + 1$ (including the original template) samples as inputs to train the neural network for 100 epochs using Adam optimizer with learning rate \texttt{lr=0.01}. We use thresholding to convert the real-valued outputs to binary vector as follows:

$$
\textbf{h}_{i}=\begin{cases}
			-1, & \text{if $\textbf{output}_{i} < 0$},\\
            +1, & \text{otherwise.}
		 \end{cases}
$$

\textbf{Loss functions:} We use three loss functions.\\
\textit{Biometric identity loss:} We use triplet margin with $\mathcal{L}_2$ distance loss with the original template as anchor ($\textbf{a}$), positive disruptors ($\textbf{p}$), negative disruptors ($\textbf{n}$), and a tunable margin parameter, $m$. 
$$L_{id}(\textbf{a},\textbf{p},\textbf{n}) = \max \{\lVert \textbf{a}-\textbf{p} \rVert_2 - \lVert \textbf{a}-\textbf{n} \rVert_2 + m, 0\}.$$ 

\noindent \textit{Binarization loss:} Note the outputs of the neural network are $[-1,+1]$. To avoid loss of information after quantization to $\{-1, +1\}$, we use MSE loss between outputs before and after binarization/thresholding as follows.
$$L_{bin}(\textbf{output},\textbf{h}) = \lVert \textbf{output}-\textbf{h} \rVert_2.$$

\noindent \textit{Diversity loss:} We want to ensure the output binary cloaked template has equal distribution of -1's and +1's such that the mean is close to zero. 
$$L_{div}(\textbf{h}, \mathbf{0}) = \lVert \textbf{h} - \mathbf{0}  \rVert_2.$$

Combining all three losses using regularization parameters $\lambda_{id} = \lambda_{bin} = \lambda_{div} = 1$, we obtain, 
$$L_{total} = \lambda_{id}L_{id}(\textbf{a},\textbf{p},\textbf{n}) + \lambda_{bin}L_{bin}(\textbf{oup},\textbf{h}) + \lambda_{div}L_{div}(\textbf{h}, \mathbf{0}).$$

\vspace{0.5pt}
\section{EXPERIMENTS AND ANALYSIS}
We conducted experiments on two datasets with two face template extractors and evaluated three metrics.

\textbf{Datasets and Face Template Extractors.} We used Labelled Faces in-the-Wild (LFW)~\cite{LFWTech} dataset in \textit{View2} protocol with 3,000 genuine and 3,000 imposter pairs. We used a subset of Celebrity Frontal-Pose Faces in-the-wild (CFP)~\cite{cfp} dataset with 1,785 genuine and 1,727 imposter pairs (selected using the protocol in~\cite{SecFace}). We used InsightFace implementation of ArcFace~\cite{arcface} and DeepFace implementation of FaceNet~\cite{Facenetimp} in our work. Hyperparameters: $m=13.0$ (ArcFace) and $m=9.0$ (FaceNet); $k=50$. FaceNet was initially designed as 128-D vector but we use its extended version FaceNet512 which is 512-D similar to ArcFace. 

\textbf{Metrics.} We assessed \textbf{biometric performance} in terms of True Match Rate (TMR) \textit{(higher is better)} at a False Match Rate (FMR)=0.001. We computed \textbf{irreversibility} in terms of Success Attack Rate (SAR) \textit{(lower is better)} while considering the full disclosure threat model (worst-case) in ISO/IEC 30136 standard, where the adversary has access to the binary cloak, the statistics of the distribution of the face templates (mean and covariance), and the trained \textsc{FaceCloak} to use it for inference. We measured \textbf{unlinkability} in terms of global system overall linkability ($D\xleftrightarrow{sys}$)~\cite{unlink}, and maximal linkability ($M\xleftrightarrow{sys}$)~\cite{maxlink} metrics. Both metrics vary from $[0,1]$, \textit{(lower is better)}, and 0 implies complete unlinkability.


\begin{table}[]
\centering
\caption{Biometric performance, irreversibility and unlinkability evaluation of \textsc{FaceCloak} on the \textbf{LFW} dataset with ArcFace and FaceNet face template extractors. The results clearly demonstrate that biometric performance loss is minimal ($<1.5\%$) and high irreversibility and unlinkability.}
\label{Tab1}
\scalebox{0.98}{
\begin{tabular}{|l||l|ll|} \hline
\textbf{Metric} & \textbf{Description }                                                                                                                                      & \textbf{ArcFace} & \textbf{FaceNet} \\ \hline
\multirow{2}{*}{{\begin{tabular}[c]{@{}l@{}}Biometric\\ Matching\end{tabular}}} & \begin{tabular}[c]{@{}l@{}}\textit{Before} \\ \textit{Cloaking}\end{tabular} & 96.2\%          & 69.3\%          \\
                                                                                                     & \begin{tabular}[c]{@{}l@{}}\textit{After}\\ \textit{Cloaking}\end{tabular}   & 95.7\%          & 67.8\%          \\ \hline \hline
{Irreversibility}                                                              & \textit{SAR}                                                      & 0.0\%            & 0.0\%            \\ \hline \hline
                                                                                                     \multirow{2}{*}{{Unlinkability}}                                                              & $D\xleftrightarrow{sys}$                                                       & 0.003            & 0.03             \\ 
                                                                                                     & $M\xleftrightarrow{sys}$                                                       & 0.007            & 0.11 \\ \hline      
\end{tabular}}
\end{table}

\textbf{Results and Discussion.} We report the \textsc{FaceCloak} results evaluation on the \textbf{LFW} dataset in Table~\ref{Tab1}. Results indicate high biometric retention and very high unlinkability. For irreversibility, we assumed that the adversary has access to the distribution of unprotected templates and sampled 10 instances from this distribution as initial guess in distinct attempts. The attacker then optimizes over the initial guess by passing it through the trained \texttt{FaceNet} and using the output to match with the binary cloak. We used Adam optimizer with \texttt{lr=0.01} and 1,000 steps to converge to $\textbf{t}^*$, and then compute its cosine similarity (inversion score) with the unprotected reference template $\textbf{t}$ and compare against the threshold @ FMR=0.001. If the inversion score exceeds the threshold in any one of the 10 attempts, we consider that as a successful attack and report the SAR as proportion of successful attacks. We present preliminary results on 500 protected templates (5K attacks); we achieve perfect irreversibility. Note our method assumes \textit{white-box} access unlike MLP-Hash~\cite{MLPhash} which uses a numerical solver and suffers from weak irreversibility. Our method is resilient due to (i) non-linear activation, (ii) unique disruptors created on the fly, (iii) feature dropout (dimensionality reduction from 512 to $d$), and (iv) binarization. On the \textbf{CFP} subset, we perform biometric matching and observe, \textit{Before cloaking} TMR=99.61\%, and \textit{After cloaking} TMR=99.23\% @FMR=0.001. Unlinkability metrics show $D\xleftrightarrow{sys}$ = 0.0005, and $M\xleftrightarrow{sys}$ = 0.003.

\textbf{Comparison to baselines.} We outperform Deep Ranking Hashing~\cite{DRHN} and IronMask~\cite{IronMask} on LFW using ArcFace in terms of biometric matching performance: TMR=88.8\%(DRH)/84.4\%(IronMask)/95.7\%\textbf{(Ours)} @FMR=0.001. We outperform MLP-Hash~\cite{MLPhash} using FaceNet on LFW: TMR= 59.4\%$\pm$5.02(MLP-Hash stolen key scenario)/67.8\%\textbf{(Ours)}. The preliminary results on the CFP subset show that our method is comparable to SecureFace~\cite{SecFace} in terms of biometric recognition and with SLERPFace~\cite{SLERP} in terms of unlinkability. \textsc{FaceCloak} is 6$\times$ faster than PolyProtect~\cite{polyprotect} in terms of matching time.

\textbf{Generalizability.} We rely on the face template extracted using existing schemes such as ArcFace and FaceNet to derive positive and negative disruptors. Therefore, the selection and creation of optimal disruptors is important for driving \textsc{FaceCloak}. So, we investigated a secondary supervision strategy that uses Orthogonal Projection Loss (OPL)~\cite{OPL} in conjunction with angular margin loss ($\mathcal{L}_{arcmargin}$) and performed preliminary experiments on the LFW dataset to test the generalizability of our method. OPL is formulated as: $\mathcal{L}_{OPL} = \bigg(1- \frac{\sum_{c_i=c_j}\langle \textbf{f}_i, \textbf{f}_j \rangle}{\sum_{c_i=c_j}\mathds{1}}\bigg) + \bigg\lvert \frac{\sum_{c_i\neq c_j}\langle \textbf{f}_i, \textbf{f}_j \rangle}{\sum_{c_i\neq c_j}\mathds{1}}\bigg\rvert.$ 
Here, $\langle \cdot, \cdot \rangle$ denotes cosine similarity operator, $\textbf{f}_i, \textbf{f}_j$ denotes $\mathcal{L}_2-$ normalized features/templates belonging to classes $c_i, c_j$, and $\lvert \cdot \rvert$ denotes absolute operator. We trained the ArcFace model (IResNet 50)~\cite{insightface} package on the VGGFace2~\cite{VGG_1} dataset ($>$200K images 8K identities) with the new loss $\mathcal{L}_{arcmargin} + \lambda_{OPL}\mathcal{L}_{OPL}$, where $\lambda_{OPL}=5$. We tested the robustness of \textsc{FaceCloak} using unprotected templates derived using $\mathcal{L}_{OPL}$ on the LFW dataset. We achieve TMR=90.1\% @ FMR=0 after cloaking and unlinkability values as $D\xleftrightarrow{sys}$=0.01 and $M\xleftrightarrow{sys}$=0.05 which shows promising results in terms of (i) capability of OPL for effective disruptor creation and (ii) robustness of \textsc{FaceCloak} towards novel loss functions for face template extraction. We will investigate its potential further in future work.

\textbf{Demographic attribute protection.} PolyProtect can leak sensitive attributes~\cite{Poly_FE} and require homomorphic encryption for user privacy. We test whether \textsc{FaceCloak}-based protected templates divulge demographic (gender) information. Note that \textsc{FaceCloak} is designed to be a template protection scheme, and not as a de-identification method for suppressing soft biometrics. We first train a 2-layer fully connected network (64$\rightarrow$32$\rightarrow$1) followed by ReLU and sigmoid activations, respectively. We train the network on a subset of cloaks generated from the LFW dataset ($\sim$3,200) and test on a disjoint set of cloaks ($\sim$1,400) such that they are evenly distributed across gender. We use DeepFace to extract ground-truth gender labels, and train the network using Adam optimization for 100 epochs with Binary Cross Entropy loss. We achieve \textbf{48.4\% gender prediction accuracy} (close to random chance=50\%), showing, our capability of of successfully suppressing sensitive demographic attributes.

\vspace{0.5pt}
\section{SUMMARY AND FUTURE WORK}
We design a neural network-driven novel face template protection scheme known as \textsc{FaceCloak} that first learns an optimal set of disruptors from a single unprotected face template and then combines them with the face template to create a binary cloak. \textsc{FaceCloak} acts as a hash function and is supervised by biometric identity, binarization and diversity losses. The novelty of our method lies in the ability to take advantage of a small set of easily regenerated disruptors that guide the network to retain face matching while mitigating inversion attacks. Experiments show that our method achieves irreversibility, biometric performance, and unlinkability/renewability while eliminating the need for secret key or user-specific parameter management, and can be easily deployed as an add-on module with minimal overhead. 

Future work will focus on evaluating our method across complex datasets, additional template extractors, modalities, and robustness to attack via record multiplicity (ARM).

\section*{ETHICAL IMPACT STATEMENT}
We have developed a face template protection scheme that protects the biometric signature of an individual by mitigating template inversion attacks, and further secures the biometric system. Our method can be used as an additional module in existing systems with little overhead. Our algorithm uses synthetic faces generated using StyleGAN3. As the generative model was trained on real faces, it may lead to inadvertent information leakage. To avoid this, we ensure that the disruptors from synthetic faces can be created on the fly. Our preliminary analysis indicates resilience against white-box attack.

\clearpage
\balance
{\small
\bibliographystyle{ieee}
\bibliography{egbib}
}

\end{document}